# How to Build an Adaptive AI Tutor for Any Course Using Knowledge Graph-Enhanced Retrieval-Augmented Generation (KG-RAG)

Chenxi Dong*
*Department of Mathematics and Information Technology*
*The Education University of Hong Kong*
Hong Kong, China
cdong@eduhk.hk
*Corresponding author

Yimin Yuan*
*Faculty of Sciences, Engineering and Technology*
*The University of Adelaide*
Adelaide, Australia
yimin.yuan@student.adelaide.edu.au
*Corresponding author

Kan Chen
*School of Communications and Information Engineering*
*Chongqing University of Posts and Telecommunications*
Chongqing, China
ck_linkin123@163.com

Shupei Cheng
*Department of Mechanical and Electronic Engineering*
*Wuhan University of Technology*
Wuhan, China
1248228520@qq.com

Chujie Wen*
*Department of Mathematics and Information Technology*
*The Education University of Hong Kong*
Hong Kong, China
cwen@eduhk.hk
*Corresponding author

*Abstract*—**Integrating Large Language Models (LLMs) in Intelligent Tutoring Systems (ITS) presents transformative opportunities for personalized education. However, current implementations face two critical challenges: maintaining factual accuracy and delivering coherent, context-aware instruction. While Retrieval-Augmented Generation (RAG) partially addresses these issues, its reliance on pure semantic similarity limits its effectiveness in educational contexts where conceptual relationships are crucial. This paper introduces Knowledge Graph-enhanced Retrieval-Augmented Generation (KG-RAG), a novel framework that integrates structured knowledge representation with context-aware retrieval to enable more effective AI tutoring. We present three key contributions: (1) a novel architecture that grounds AI responses in structured domain knowledge, (2) empirical validation through controlled experiments (n=76) demonstrating significant learning improvements (35% increase in assessment scores, p<0.001), and (3) a comprehensive implementation framework addressing practical deployment considerations. These results establish KG-RAG as a robust solution for developing adaptable AI tutoring systems across diverse educational contexts.**

## I. INTRODUCTION

Intelligent Tutoring Systems (ITS) enable students to learn independently. However, it is hindered by two key challenges: (1) Information Hallucination: Large Language Models (LLMs), the engines behind many ITS, can generate plausible but factually incorrect information [4]. (2) Conceptual coherence in response: Existing AI tools often struggle to understand the interconnected relationships between course concepts, limiting their ability to deliver coherent responses [3]. Approaches like retrieval-augmented generation (RAG) rely on semantic similarity and cannot capture the structured relationships between concepts. We introduce KG-RAG (Knowledge Graph-enhanced Retrieval-Augmented Generation) to overcome these limitations. This novel approach integrates structured knowledge graphs to enhance the coherence of responses, as demonstrated by a Q&A example in Fig. 1.

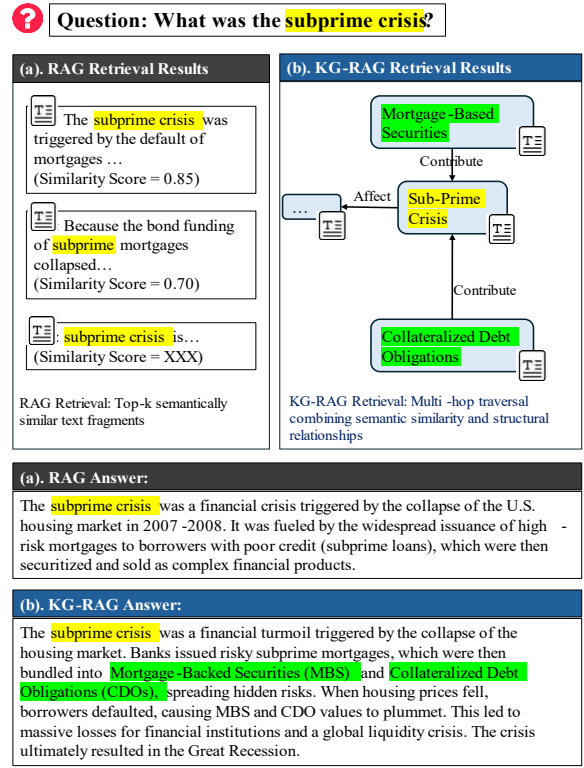

Fig. 1. KG-RAG: Using Structured Knowledge Retrieval to Enhance Response Quality

Fig 1 illustrates the key advantage of our KG-RAG approach in enhancing contextual understanding. For a finance course, when a student queries the subprime crisis, standard RAG (Fig. 1a) retrieves isolated text snippets, potentially leading to incomplete and disconnected answers. In contrast, KG-RAG (Fig. 1b) significantly enhances retrieval by explicitly structuring relationships between related concepts, such as MBS (Mortgage-Backed Securities) and CDOs (Collateralized Debt Obligations). This structured retrieval enables AI tutors to maintain a logical conceptual flow, mirroring expert pedagogical strategies. Key contributions of our work include:

1) **Novel KG-RAG Architecture for Enhanced Answer Completeness**: We present KG-RAG, a framework that enhances AI tutoring through structured knowledge integration.

2) **Empirical Validations**: We present comprehensive evaluations through controlled experiments and user studies (n=76) to demonstrate the efficacy of our approach.

3) **Practical Deployment Framework and Cost Analysis**: We provide a thorough discussion on cost analysis, global accessibility, and data privacy strategies, making the framework suitable for real-world deployment.

The remainder of this paper is organized as follows: Section II presents the foundations of our framework. Section III details the KG-RAG system architecture. Section IV describes our results. Section V discusses practical considerations and limitations. Finally, Section VI concludes the paper.

## II. FOUNDATIONS OF KG-RAG

Our KG-RAG framework is built upon three core technologies: Large Language Models (LLMs), Retrieval-Augmented Generation (RAG), and Knowledge Graphs (KG).

### A. LLMs: Powering Language Generation

Large Language Models (LLMs), based on deep neural networks trained on extensive text corpora, have revolutionized natural language processing by utilizing transformers with self-attention to capture contextual information [10]. Large Language Models serve as our system's primary reasoning and generation engine, and we employ DeepSeek-V3, a state-of-the-art Mixture-of-Experts (MoE) model, in this work. DeepSeek-V3 excels in domains requiring complex reasoning, demonstrated by its 90.2% accuracy on the MATH benchmark, outperforming other advanced models like GPT-4 and Claude 3 Opus [11]. The selection of DeepSeek-V3 is motivated by its:

1) **Global accessibility:** Unlike many other models, DeepSeek-V3 is open-source and free from regional restrictions (e.g., OpenAI's service limitations in certain regions like China).

2) **Superior Reasoning Ability:** DeepSeek-V3 excels in reasoning tasks, which is vital for the AI tutor.

3) **Cost Efficiency:** Using DeepSeek-V3 through the Alibaba Cloud API reduces operational costs to approximately 7% of GPT-4's expenses, making it a viable choice for large-scale educational applications.

In its basic implementation, the LLM generates a response (r) solely based on the input query (q):

$$r_{LLM} = LLM(q) \quad (1)$$

### B. RAG: Enhancing Accuracy and Relevance

Retrieval-augmented generation (RAG) addresses the issue of information hallucination by grounding the responses of LLMs in verified source materials [12]. As shown in Fig. 2, RAG embeds the queries and knowledge base documents into a shared vector space, allowing for similarity-based retrieval of relevant context.

**Retrieval-Augmented Generation (RAG)**

**Input**: Query q, Knowledge base KB={$d_1,…,d_N$}, k=5

**Output**: Response r

1. Compute embeddings:
   1. $e_q \leftarrow$ Embed (q)
   2. E ← {Embed (di) | di∈KB}
2. Compute similarities: S ← {cos ($e_q, e_{di}$) | $e_{di} \in E$}
3. Retrieve top-k similar documents indices: $I_{top} \leftarrow \underset{k}{\arg\max} S$
(returns indices of k highest scores)
4. Generate context: c ← concatenate ({di| i∈ $I_{top}$})
(combines selected documents into a single text)
5. Generate response: r ← LLM (c,q) (large language model generates answer using context and query)
6. **Return** r

Fig. 2. Retrieval-Augmented Generation (RAG) Workflow.

Using RAG, the LLM generates the response by integrating this similarity-based context ($C_{similarity}$) with the query (q):

$$r_{RAG} = LLM(C_{similarity}, q) \quad (2)$$

### C. Knowledge Graph for Contextual Enhancement

Knowledge Graphs (KGs) enhance the AI Tutor's contextual understanding by structuring domain knowledge into interconnected nodes and edges, aligning with constructivist learning theory, where learners build knowledge through relationship formation [6]. We constructed a knowledge graph from course materials using DeepSeek-V3, where KG = (V, E). The graph's vertices (V) represent the core domain concept. For example, a finance course might use a knowledge graph where security types such as "Mortgage-Backed Securities" (MBS) and events like the "Subprime Crisis" are represented as nodes, and the relationships between them (e.g., "MBS are affected by Subprime Crisis") are captured as edges (E). We apply Knowledge-Guided Retrieval (KGR) to extract expanded context $C_{expanded}$ by traversing the knowledge graph (KG):

$$C_{expanded} = KGR(KG, q) \quad (3)$$

This retrieval process ensures that the LLM receives richer, interconnected context, which leads to more contextually aware responses. The final KG-RAG response integrates both similarity-based and knowledge-guided context:

$$r_{KG-RAG} = LLM(C_{similarity}, C_{expanded}, q) \quad (4)$$

This KG structured representation resolves context fragmentation by enabling knowledge-guided retrieval and enhances the reasoning process's explainability.

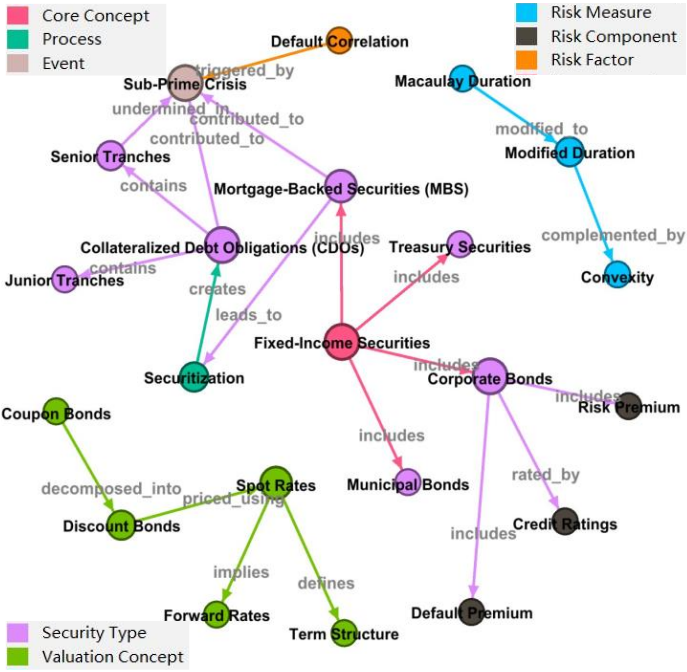

Fig. 3. Knowledge Graph Extracted from a Lecture Notes (https://ocw.mit.edu/courses/15-401-finance-theory-i-fall-2008/resources/mit15_401f08_lec04/).

### III. AI TUTOR DESIGN: THE KG-RAG SOLUTION

Our AI Tutor implementation integrates LLM (DeepSeek-V3), RAG, and KG into a unified framework, delivering precise, context-aware educational responses, as shown in Fig. 4. The system comprises three main steps.

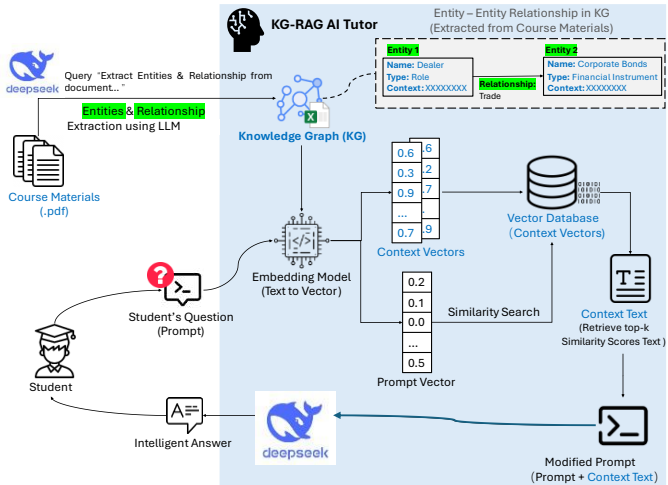

Fig. 4. Architecture of the KG-RAG AI Tutor.

#### A. KG Construction and Validation

The knowledge graph construction process combines automated extraction with expert validation to ensure efficiency and accuracy. We first segment course materials into 1,000-token chunks and process them using DeepSeek-V3 with standardized prompts (e.g., "Extract entities and relationships from the following text in [Entity1, Relationship, Entity2] format"). This automated extraction yields structured triples (e.g., [Mortgage-Backed Securities, Affects, Sub-Prime Crisis]) stored in a .xlsx file. A domain expert, a finance lecturer, validates each extracted triple for precision (is the relationship correct?), completeness (are key relationships captured?), and pedagogical relevance (is the relationship meaningful for learning the material?). This two-stage approach ensures that the resulting knowledge graph accurately represents the content structure.

#### B. Knowledge-Guided Retrieval (KGR)

KGR is the core innovation of KG-RAG. KGR first converts student queries into semantic vectors using text-embedding models (text-embedding-v2 from Alibaba Cloud). Unlike traditional RAG, which retrieves isolated text snippets, KGR expands the retrieval process by traversing the knowledge graph to incorporate related concepts (Shown in Fig. 5). For example, a query about "mortgage-backed securities" triggers the retrieval not only of directly related content but also of connected concepts such as "subprime crisis" and "fixed-income securities." The KGR ensures that the LLM receives a richly interconnected context to generate its response.

---

**Knowledge-Guided Retrieval (KGR)**

**Input**:
- Query q;
- Knowledge graph KG=(V, E)
  - V: Nodes (entities), each with a context attribute (textual content).
  - E: Edges (relationships between nodes).
- k=5: Number of top nodes to retrieve initially
- d=max: Traversal depth (retrieve all connected nodes).

**Output**: Expanded context $C_{expanded}$

1. Compute node similarities:
   1. Embed query $e_q \leftarrow$ Embed (q).
   2. For each node $V_i \in V$, compute embedding $e_{vi} \leftarrow$ Embed (context ($v_i$))
   3. Calculate similarity scores: $S \leftarrow \{s_i = \cos(e_q, e_{vi}) \mid v_i \in V\}$.

2. Retrieve top-k nodes: $V_{top} \leftarrow \underset{k}{\arg\max} \, S$.

3. Traverse all connected nodes: $V_{traversed} \leftarrow$ TraverseKG (v, d=max)

TraverseKG (v,d): Recursively collects all nodes connected to $v \in V_{top}$ via edges E across the KG.

4. Generate expanded context ($C_{expanded}$):
$C_{expanded} \leftarrow$ concatenate({context($V_i$) | $V_i \in V_{traversed}$})

**5. Return $C_{expanded}$**.

Fig. 5. Knowledge-Guided Retrieval (KGR) Workflow.

#### C. Response Synthesis

Building on the context from KGR, the response synthesis leverages DeepSeek-V3 to generate the final answer. Using designed prompts - "You are an expert tutor. Using the following course material: [retrieved content], please answer the student's question: [original query]. Explain concepts clearly with detail."—The system guides the LLM in producing relevant and complete responses.

### IV. AI TUTOR EVALUATION

Our evaluation includes (a) qualitative response analysis, (b) student feedback assessment, and (c) a controlled experimental study.

## A. Comparative Analysis of Response Quality

Table I presents representative examples highlighting the improvements achieved by integrating knowledge graphs into the response generation process.

TABLE I. COMPARISON OF RESPONSES FROM RAG AND KG-RAG

| **Q1. What is Treasury Security, and what are its main characteristics?** | |
|---|---|
| **RAG Answer:** Treasury securities are financial instruments issued by the U.S. government. They include bills (short-term), notes, and bonds (longer-term). They are considered low risk due to the government's credit. | **AI Tutor Answer (KG-RAG):** Treasury Securities are debt the U.S. government issues. They come in three main types: Treasury Bills (maturity < 1 year, no coupons), Treasury Notes (2-10 years, semi-annual coupons), and Treasury Bonds (20-30 years, semi-annual coupons). They are the safest fixed income, backed by the U.S. government's credit. (KG-RAG Improvement in Completeness) |
| **Q2. Explain what duration means in the context of fixed-income securities.** | |
| **RAG Answer:** Duration measures a bond's price sensitivity to changes in interest rates, representing the weighted average time until a bond's cash flows are received. It quantifies interest rate risk. | **AI Tutor Answer (KG-RAG):** Duration measures a bond's price sensitivity to yield changes, approximating the percentage price change for a 1% yield shift. Modified duration, which equals Macaulay duration divided by (1 + yield), is a more commonly used metric. (KG-RAG Improvement in: Precision) |
| **Q3. What are the connections between MBS, CDOs, and Sub-Prime Crisis?** | |
| **RAG Answer:** Mortgage-backed securities (MBS) and CDOs were pivotal in the subprime crisis, as high-risk loans were repackaged into misjudged securities. Poor underlying loan quality and decreased housing values led to massive defaults and losses, particularly affecting lower traces of products like CDOs. | **AI Tutor Answer (KG-RAG):** The subprime crisis revealed how risks propagated through connections between Mortgage-Backed Securities (MBS) and Collateralized Debt Obligations (CDOs): 1) Defaults on subprime mortgages disrupted cash flows to MBS, starting the chain reaction, 2) CDOs repackaged MBS risk into seemingly safer tranches, 3) Default correlations were underestimated, 4) Rating downgrades triggered forced selling, 5) Market illiquidity amplified losses. (KG-RAG Improvement in: Relationship Preservation) |

The comparison reveals three key advantages of the KG-RAG system:

1) Completeness: In Q1, both systems correctly identify Treasury securities as U.S. debt instruments. However, KG-RAG provides a more thorough explanation, specifying the types (Bills, Notes, and Bonds), their maturity, and payment structure.

2) Precision: KG-RAG offers more precise explanations of technical concepts by adding crucial context explaining the calculation and application of modified duration that is absent in the RAG response.

3) Relationship Preservation: In Q3, KG-RAG delineates the relationships between MBS, CDOs, and the subprime crisis, presenting them in a coherent, step-by-step narrative.

## B. Student Feedback Analysis

Student feedback was collected from 76 university participants using a 5-point Likert scale. Table II summarizes that statistical analysis via one-sample t-tests (comparing against a neutral midpoint of 3.0) revealed significant positive shifts across several dimensions.

TABLE II. STATISTICS OF STUDENTS FEEDBACK (N=76)

| Evaluation Criterion | Response Distribution (%) | Mean | SD | p* |
|---|---|---|---|---|
| 1. The Answer is Relevant | 0% (Strongly Disagree), 0% (Disagree), 16% (Neutral), **50% (Agree), 34% (Strongly Agree)** | 4.18 | 0.78 | <.001 |
| 2. The Answer is Detailed | 6% (Strongly Disagree), 16% (Disagree), 40% (Neutral), **20% (Agree), 18% (Strongly Agree)** | 3.28 | 1.15 | 0.289 |
| 3. My Understanding Improved | 20% (Strongly Disagree), 4% (Disagree), 14% (Neutral), **38% (Agree), 24% (Strongly Agree)** | 3.42 | 1.02 | 0.003 |
| 4. Easy to use | 0% (Strongly Disagree), 14% (Disagree), 40% (Neutral), **10% (Agree), 36% (Strongly Agree)** | 3.68 | 0.81 | <.001 |
| 5. Better than a Human Tutor | 6% (Strongly Disagree), 8% (Disagree), 26% (Neutral), **24% (Agree), 35% (Strongly Agree)** | 3.71 | 1.08 | <.001 |

* p < 0.05 indicates statistical significance.

Response relevance emerged as a particular strength, with 84% of participants indicating favorable agreement (M = 4.18, SD = 0.78, p < .001). The system also demonstrated strong accessibility (M = 3.68, SD = 0.81, p < .001) and compared favorably to traditional human tutoring (M = 3.71, SD = 1.08, p < .001). Student understanding showed significant improvement (M = 3.42, SD = 1.02, p = 0.003), though response detail (M = 3.28, SD = 1.15, p = 0.289) emerged as an area for potential enhancement.

## C. Controlled Experiment: KG-RAG vs. Standard RAG

We conducted a controlled experiment comparing KG-RAG against standard RAG-based tutoring. The study involved 76 first-year university students (36 female, 40 male) with no prior finance coursework, randomly assigned to either the Control Group (standard RAG, n=38) or the Experimental Group (KG-RAG, n=38). The experimental protocol consisted of:

1) A standardized 3-hour self-study period using course materials (https://ocw.mit.edu/courses/15-401-finance-theory-i-fall-2008/resources/mit15_401f08_lec04/).

2) A 45-minute interactive session with the assigned tutoring system (RAG vs. KG-RAG)

3) A multiple-choice assessment (10-point scale) drafted by a domain expert (https://github.com/098765d/KG-RAG/blob/f5b4fed409af6661aabe70a3dd73c101625423fd/MC_quiz.pdf). The results are shown in Table III.

The assessment results (Table III) demonstrate significant performance differences between the two groups. The KG-RAG group achieved substantially higher scores (M = 6.37, SD = 1.92) than the RAG group (M = 4.71, SD = 1.93). Statistical analysis confirms the significance of this difference (t = -3.75, p = 0.00035) with a large effect size (Cohen's d = 0.86). This performance differential aligns with constructivist learning theory, suggesting that KG-RAG's ability to present interconnected concepts in a structured manner facilitates more effective knowledge integration. The significant effect size indicates that these improvements are statistically significant and practically meaningful for educational outcomes

TABLE III. ASSESSMENT RESULTS (N=76, 10-POINT SCALE)

| Group | Mean | SD | Statistical Comparison | Effect Size |
|---|---|---|---|---|
| Control (RAG) | 4.71 | 1.93 | t=−3.75, p=.00035 | Cohen's d=0.86 |
| Experimental (KG-RAG) | 6.37 | 1.92 | | |

## V. DISCUSSIONS

This section discusses practical considerations that influence real-world deployment, including economic feasibility, data privacy, and system limitations.

### A. Economic Feasibility

The economic sustainability of AI tutoring systems represents a critical consideration for educational institutions, particularly those operating under resource constraints. Our comparative analysis of LLM operational costs (Table IV) unequivocally positions DeepSeek-V3 as an economically advantageous choice. The order-of-magnitude cost difference compared to models like GPT-o1 translates into substantial savings for institutions. For instance, a university supporting just 1,000 students could realize annual savings nearing $100,000 by opting for DeepSeek-V3 over GPT-o1. This economic advantage is not merely about budget savings but democratizing access to high-quality, personalized tutoring. By minimizing operational costs, KG-RAG powered by DeepSeek-V3 becomes a viable option for institutions globally, including those in underserved communities where access to individualized educational support is often limited.

TABLE IV. OPERATIONAL COST COMPARISON

| LLM Options | Cost Per Q&A (USD) | Cost Per Q&A (CNY) | Cost Ratio vs. DeepSeek-V3 |
|---|---|---|---|
| GPT-o1 (OpenAI - US) | 2.98e-3 | 2.19e-2 | 13.7× |
| Qwen-2.5-72b (Alibaba Cloud - CN) | 3.27e-4 | 2.40e-3 | 1.5× |
| DeepSeek-V3 (DeepSeek AI - CN) | 2.18e-4 | 1.60e-3 | 1.0 |

To optimize operational costs, we propose a chat history re-usage mechanism (Fig. 6). This involves comparing the vector embeddings of a new user query with those from previous queries stored in the chat history, using cosine similarity. An open-source embedding model (e.g., MiniLM-L6-V2) can be used for this comparison. If the similarity score exceeds a predefined threshold (e.g., 0.85), the relevant answer is retrieved directly from the chat history, avoiding reprocessing the query.

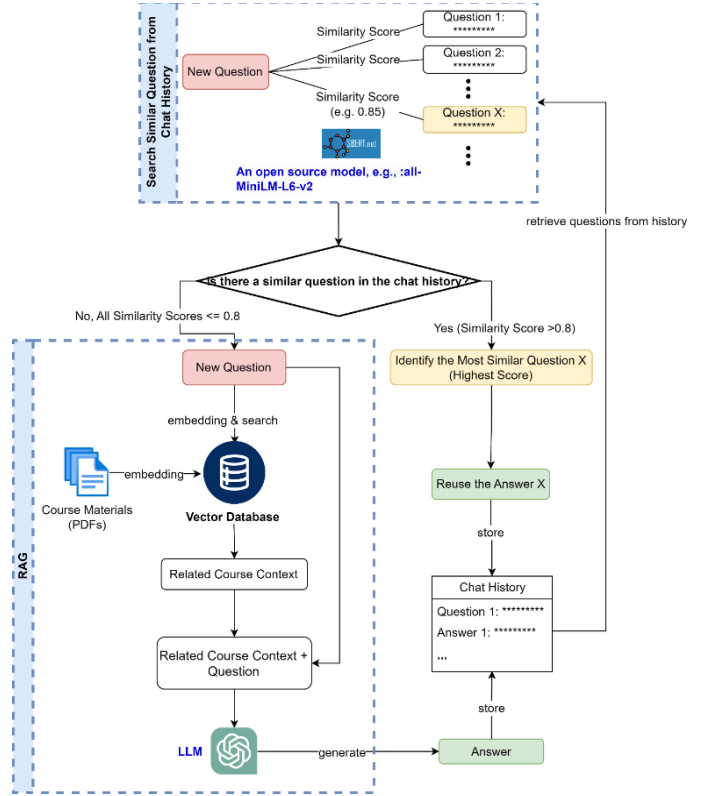

Fig. 6. Workflow for Efficient Question Answering via Chat History Reusage.

### B. Global Accessibility

The geopolitical limitations of certain prominent LLMs (e.g., OpenAI service regional restrictions) highlight a critical challenge for global AI equity. KG-RAG's implementation with DeepSeek-V3, a globally accessible and open-source alternative, addresses this concern. This vendor diversification is not merely a technical choice; it is an ethical imperative to ensure that the benefits of AI-driven education are not geographically restricted. By embracing globally accessible technologies, we can actively work towards bridging the digital divide in AI education, ensuring that students worldwide, regardless of their location or institutional affiliations, can benefit from advanced tutoring systems like KG-RAG. Further exploration of regionally developed and culturally adapted knowledge graphs could also enhance the relevance and effectiveness of KG-RAG across diverse educational contexts.

### C. Data Privacy

Given the sensitivity of educational data, local deployment on dedicated servers or private clouds is recommended. This approach minimizes data exposure and aligns with institutional privacy policies and regulatory requirements.

### D. Limitations and Future Directions

While this study demonstrates the potential of the KG-RAG framework for AI tutoring, several limitations suggest important avenues for future work. One notable limitation lies in the scale and scope of our user study, which involved a relatively small sample (n=76) of university students. While the results indicate the system's effectiveness in enhancing learning outcomes, more extensive and diverse participant groups—potentially spanning multiple institutions and academic disciplines—would provide more substantial evidence of its generalizability. A broader study could also capture a wider range of user needs, language proficiencies, and cultural contexts, illuminating how these variables might impact user engagement and knowledge retention over extended periods. Second, the present evaluation relies primarily on a qualitative assessment. A promising direction for future work would be to complement these qualitative findings with quantitative text quality metrics, such as BLEU or ROUGE [16].

## VI. Conclusion

This paper has presented KG-RAG, an innovative framework that addresses fundamental challenges in AI tutoring by integrating structured knowledge representations with advanced language models. Our evaluation demonstrates significant improvements in learning outcomes, with KG-RAG substantially outperforming traditional RAG-based approaches in both quantitative metrics and qualitative assessments. Furthermore, our implementation using DeepSeek-V3, a globally accessible and cost-effective LLM from Alibaba Cloud, directly addresses the over-reliance on US-based models and promotes a more inclusive landscape for AI in education.

Future research will focus on validating KG-RAG across diverse academic domains and learner populations, incorporating objective text quality metrics, automating knowledge graph construction, and enhancing adaptive scaffolding capabilities for a more profound learning experience. Thus, KG-RAG offers a robust and economically viable blueprint for developing AI tutors, paving the way for effective and globally accessible customized learning solutions.

## References

[1] A. O. Tzirides et al., "Combining human and artificial intelligence for enhanced AI literacy in higher education," Computers and Education Open, vol. 6, 2024, Art. no. 100184.

[2] Baumgart, A., & Madany Mamlouk, A. (2022). A Knowledge-Model for AI-Driven Tutoring Systems. In *Information Modelling and Knowledge Bases XXXIII* (pp. 1-18). IOS Press.

[3] Bui, T., Tran, O., Nguyen, P., Ho, B., Nguyen, L., Bui, T., & Quan, T. (2024, June). Cross-Data Knowledge Graph Construction for LLM-enabled Educational Question-Answering System: A Case Study at HCMUT. In *Proceedings of the 1st ACM Workshop on AI-Powered Q&A Systems for Multimedia* (pp. 36-43).

[4] Cai, D., Wang, Y., Liu, L., & Shi, S. (2022, July). Recent advances in retrieval-augmented text generation. In Proceedings of the 45th International ACM SIGIR Conference on Research and Development in Information Retrieval (pp. 3417-3419).

[5] Devlin, J., Chang, M. W., Lee, K., & Toutanova, K. (2018). Bert: Pre-training of deep bidirectional transformers for language understanding. arXiv preprint arXiv:1810.04805.

[6] Edge, D., Trinh, H., Cheng, N., Bradley, J., Chao, A., Mody, A., ... & Larson, J. (2024). From local to global: A graph rag approach to query-focused summarization. *arXiv preprint arXiv:2404.16130*.

[7] GLM, T., Zeng, A., Xu, B., Wang, B., Zhang, C., Yin, D., ... & Wang, Z. (2024). Chatglm: A family of large language models from glm-130b to glm-4 all tools. *arXiv preprint arXiv:2406.12793*.

[8] Gromyko, V. I., Kazaryan, V. P., Vasilyev, N. S., Simakin, A. G., & Anosov, S. S. (2017, August). Artificial intelligence as tutoring partner for human intellect. In International Conference of Artificial Intelligence, Medical Engineering, Education (pp. 238-247). Cham: Springer International Publishing.

[9] Kasneci, E., Seßler, K., Küchemann, S., Bannert, M., Dementieva, D., Fischer, F., ... & Kasneci, G. (2023). ChatGPT for good? On opportunities and challenges of large language models for education. Learning and individual differences, 103, 102274.

[10] Lewis, P., Perez, E., Piktus, A., Petroni, F., Karpukhin, V., Goyal, N., ... & Kiela, D. (2020). Retrieval-augmented generation for knowledge-intensive nlp tasks. Advances in Neural Information Processing Systems, 33, 9459-9474.

[11] Liu, A., Feng, B., Xue, B., Wang, B., Wu, B., Lu, C., ... & Piao, Y. (2024). Deepseek-v3 technical report. *arXiv preprint arXiv:2412.19437*.

[12] L. Guo et al., "Evolution and trends in intelligent tutoring systems research: a multidisciplinary and scientometric view," Asia Pacific Education Review, vol. 22, pp. 441-461, 2021.

[13] Moore, S., Tong, R., Singh, A., Liu, Z., Hu, X., Lu, Y., … & Stamper, J. (2023, June). Empowering education with llms-the next-gen interface and content generation. In International Conference on Artificial Intelligence in Education (pp. 32-37). Cham: Springer Nature Switzerland. https://doi.org/10.1007/978-3-030-78270-2_6

[14] Nye, B., Mee, D., & Core, M. G. (2023). Generative large language models for dialog-based tutoring: An early consideration of opportunities and concerns. In AIED Workshops.

[15] OpenAI, R. (2023). Gpt-4 technical report. arxiv 2303.08774. View in Article, 2, 13.

[16] Sarmah, B., Mehta, D., Hall, B., Rao, R., Patel, S., & Pasquali, S. (2024, November). Hybridrag: Integrating knowledge graphs and vector retrieval augmented generation for efficient information extraction. In *Proceedings of the 5th ACM International Conference on AI in Finance* (pp. 608-616).

[17] Y. Lee, "Developing a computer-based tutor utilizing Generative Artificial Intelligence (GAI) and Retrieval-Augmented Generation (RAG)," Education and Information Technologies, 2024.

[18] Yang, A., Yang, B., Zhang, B., Hui, B., Zheng, B., Yu, B., ... & Qiu, Z. (2024). Qwen2. 5 Technical Report. *arXiv preprint arXiv:2412.15115*.

[19] Yang, Z., Wang, Y., Gan, J., Li, H., & Lei, N. (2021). Design and research of intelligent question-answering (Q&A) system based on high school course knowledge graph. *Mobile Networks and Applications*, 1-7.